\pdfoutput=1
\documentclass{article}

\PassOptionsToPackage{numbers, compress}{natbib}

\usepackage[dblblindworkshop, final]{neurips_2026}

\usepackage[utf8]{inputenc} 
\usepackage[T1]{fontenc}    
\usepackage{hyperref}       
\usepackage{url}            
\usepackage{booktabs}       
\usepackage{amsfonts}       
\usepackage{amssymb}        
\usepackage{nicefrac}       
\usepackage{microtype}      
\usepackage{xcolor}         

\usepackage{graphicx} 
\usepackage{wrapfig}  
\usepackage{amsmath}  
\usepackage{bbm}
\usepackage[math]{cellspace}
\usepackage{mathtools}
\usepackage[capitalize,noabbrev]{cleveref} 
\usepackage{subcaption} 
\usepackage{float}
\usepackage{array}
\usepackage{lipsum} 
\usepackage{cprotect}
\usepackage{verbatim}
\usepackage{listings}
\usepackage{multirow}
\usepackage[euler]{textgreek}
\usepackage{pmboxdraw}
\usepackage{tcolorbox}

\usepackage{colortbl}  
\usepackage{makecell}  
\newcommand{\swatch}[2]{{\setlength{\fboxsep}{1.2pt}\colorbox{#1}{#2}}}  
\newcommand{\hd}[1]{{\footnotesize #1}}  
\newcommand{\sdsub}[1]{_{\text{\tiny$\pm$#1}}}
\newcommand{\nc}[2]{$#1\sdsub{#2}$}
\newcommand{\dcell}[3]{\ifdim#2pt>1.99pt\cellcolor{teal!18}\else\ifdim#2pt<-1.99pt\cellcolor{red!15}\fi\fi$#1{#2}\sdsub{#3}$}
\newcommand{\dd}[2]{\dcell{}{#1}{#2}}
\newcommand{\bd}[2]{\dcell{\mathbf}{#1}{#2}}
\newcommand{\stk}[2]{\makecell{#1\\[-2.5pt]{\tiny$\pm$#2}}}
\newcommand{\lshade}[1]{\ifdim#1pt>1.99pt\cellcolor{teal!18}\else\ifdim#1pt<-1.99pt\cellcolor{red!15}\fi\fi}
\newcommand{\lnc}[2]{\stk{$#1$}{#2}}

\usepackage{amsmath,amsfonts,bm}
\usepackage{mathtools} %

\def\eqref#1{equation~\ref{#1}}

\def\1{\bm{1}}

\def\vx{{\bm{x}}}

\DeclareMathAlphabet{\mathsfit}{\encodingdefault}{\sfdefault}{m}{sl}
\SetMathAlphabet{\mathsfit}{bold}{\encodingdefault}{\sfdefault}{bx}{n}

\newcommand\normtwo[1]{\lVert#1\rVert_{2}}

\title{On the Limits of Representation-Based \\Test-Time Adaptation for Multimodal LLMs}
\workshoptitle{VLM4RWD}

\author{
  Mikołaj Piórczyński\textsuperscript{1,2} \And
  Wojciech Masarczyk\textsuperscript{1} \And
  Damian Sójka\textsuperscript{3, 4} \AND
  Sebastian Cygert\textsuperscript{5, 6} \And
  Bartłomiej Twardowski\textsuperscript{1, 7}
}    

\begin{document}

\maketitle

\begingroup
\renewcommand{\thefootnote}{}
\footnotetext{
  \textsuperscript{1}IDEAS Research Institute, 
  \textsuperscript{2}Warsaw University of Technology, 
  \textsuperscript{3}Poznań University of Technology, 
  \textsuperscript{4}Akces NCBR,
  \textsuperscript{5}Gdańsk University of Technology,
  \textsuperscript{6}NASK National Research Institute,
  \textsuperscript{7}Computer Vision Center, Barcelona
}
\endgroup

\begin{abstract}
Vision-language models suffer significant accuracy drops under common corruptions. While test-time adaptation (TTA) successfully mitigates these shifts in CLIP-like contrastive models, its application to generative multimodal large language models (MLLMs) remains underexplored and uniquely challenging due to their sequential token generation. 
In this work, we demonstrate that representation-based TTA fails to recover performance in MLLMs. Our experiments reveal that even when violating standard TTA assumptions by leveraging target data offline, sample-blind feature alignment recovers only marginal accuracy. We attribute this failure to the nature of corruption shifts: unlike CLIP-like architectures, which benefit from a shared domain-shift direction within a single pooled classification embedding, MLLM visual token sequence shifts are nearly orthogonal and span the full embedding space. 
These findings bound what any sample-blind, position-shared correction at a single encoder site can achieve,
establishing that representation-space interventions alone might be insufficient to ensure corruption robustness in MLLMs.

\end{abstract}

\section{Introduction}
\vspace{-4pt}

\begin{wrapfigure}[18]{r}{0.43\linewidth}
    \vspace{-10pt}
    \centering
    \includegraphics[width=0.95\linewidth]{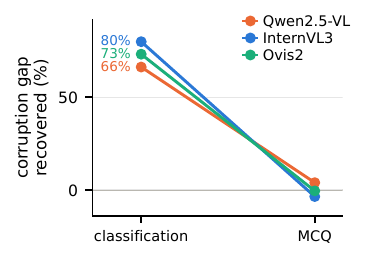}
    \vspace{-4pt}
    \caption{\textbf{Matching corrupted activations to clean statistics repairs a linear probe, but not the MLLM's answers.} For the probe, the statistics are matched on one pooled vector per image. For the MLLM, they are matched on individual~visual~tokens.}
    \label{fig:teaser}
    \vspace{-10pt}
\end{wrapfigure}

Vision-language models (VLMs) excel on a broad range of visual tasks, yet they inherit the brittleness of their visual encoders. 
Common corruptions such as sensor noise, defocus, and brightness shifts can substantially degrade accuracy even when the scene remains clearly recognizable. 
Since this degradation arises at deployment, where only unlabeled, already-corrupted inputs are available, test-time adaptation (TTA) fits precisely this setting, adapting a pretrained model to the incoming test stream without labels.

VLMs, however, span substantially different architectures. 
Contrastive models such as CLIP~\citep{radford2021clip} and SigLIP~\citep{zhai2023siglip} align images and text through a single pooled embedding per modality, whereas multimodal large language models (MLLMs) pass a sequence of visual tokens through a connector into an autoregressive language model that generates the response. 
Existing TTA work for VLMs targets almost exclusively the former: the recent TTA-VLM benchmark evaluates fifteen methods across fifteen datasets, finds gains over the earliest baseline to be marginal even there, and leaves evaluation beyond CLIP-like models and beyond classification as open problems~\citep{sheng2026illusion}. 
To our knowledge, adapting MLLMs to common corruptions at test time has not been studied, leaving practitioners who deploy them in noisy real-world environments with little evidence about whether their degraded visual performance can be recovered at all.

A natural starting point for recovering it is representation-based TTA, which corrects internal activations at inference time without updating any weights. 
If a corruption induces a consistent shift in representation space, a single sample-blind correction, estimated from activation statistics and applied to every input, should in principle undo it. 
Such methods need no gradients or labels and assume no fixed output space, making them among the few TTA approaches that remain practical at MLLM scale (\Cref{sec:related_works}).
They have been validated, however, mainly on classification, where a prediction follows from one pooled representation per image. 
The token-sequence setting is substantially more demanding: a successful correction must restore the structure of the entire sequence rather than a single image-level vector.

We therefore ask whether sample-blind representation corrections can recover the performance MLLMs lose to common corruptions. 
We apply feature alignment, matching the per-dimension mean and variance of visual-token activations to their clean values, at layers spanning the vision encoders of three modern MLLMs on RealWorldQA under 15 common corruptions. 
Accuracy stays close to the no-TTA baseline throughout: no layer, model, or corruption closes a meaningful fraction of the corruption gap, even when the alignment maps are fitted offline from paired clean--corrupted activations (\Cref{fig:upper_bound}). 
The contrast in \Cref{fig:teaser} is the central one: aligning corrupted projector activations with the clean distribution recovers most of the gap for a linear probe on a pooled representation, but none of it for the MLLM's own answers.

Why these corrections fail is a property of the shifts themselves. 
They are nearly orthogonal across images and collectively span almost the full embedding space (\Cref{fig:cos_rank}), so the component a sample-blind correction can remove accounts for only a small fraction of the total (\Cref{fig:decomposition}). 
Nor does any small subset of patches or feature dimensions carry a disproportionate share of the effect (\Cref{fig:substitute_neurons_patches_qwen}). 
There is no shared direction to remove: effective restoration would have to be specific not only to each sample but to each token within it (\Cref{sec:analysis}).

Our contributions are as follows:
\vspace{-4pt}
\begin{itemize}
    \item We provide, to our knowledge, the first evaluation of test-time adaptation for MLLMs, studying representation-based corrections across three models and 15 corruptions. We find no practically significant gains.
    \item We characterize the corruption-induced shifts and explain why sample-blind corrections fail, including a quantitative bound on the fraction of the shift that such corrections can remove.
\end{itemize}
\vspace{-4pt}

\vspace{-6pt}
\section{Related Work}
\label{sec:related_works}
\vspace{-4pt}

\vspace{-4pt}
\paragraph{Robustness of Vision-Language Models}

Despite their impressive capabilities, VLMs exhibit well-documented shortcomings: they hallucinate content not grounded in the visual input~\citep{liu2025reducing}, overlook information present in their own visual representations in favor of language priors~\citep{tong2024eyes, fu2025hidden, vo2026vision}, fail to express reduced confidence on ambiguous inputs~\citep{wang2025vision}, and flip answers under minor input perturbations~\citep{rosenfeld2025questioning}.

A complementary line of work probes this fragility with common image corruptions, popularized by ImageNet-C~\citep{hendrycks2019imagenet_c} as a controlled and reproducible testbed for the degradations encountered in real-world deployment; a growing body of benchmarks extends this paradigm to MLLMs~\citep{chen2023benchmarking, li2024rbench, qiu2025benchmarking, saxena2026vlmrobustbench}.
Notably, despite internet-scale pretraining, MLLMs still degrade substantially under such corruptions regardless of model scale, yet these benchmarks diagnose the problem without proposing remedies.

Among the few mitigation approaches, \citet{liu2025reducing} steer latent representations at inference time to reduce hallucinations, while Manifold-Adversarial Adapters (MAA)~\citep{li2026robust} improve corruption robustness by aligning features of corrupted images with their clean counterparts. MAA, however, requires a large training set of paired clean--corrupted images and multiple epochs of adapter training--assumptions violated in deployment, where only unlabeled, already-corrupted test samples are available.

To the best of our knowledge, no prior work adapts general-purpose MLLMs to corrupted images directly at test time, using only unlabeled test samples--the gap this work addresses.

\vspace{-4pt}
\paragraph{Test-Time Adaptation}

Test-Time Adaptation (TTA) addresses distribution shift at deployment by adapting a pretrained model to an unlabeled test stream~\citep{xiao2024beyond, liang2025comprehensive}.
Classic approaches utilize backpropagation-based weight updates with self-supervised objectives that fall into three families: entropy minimization, as in TENT~\citep{wang2020tent} and SAR~\citep{niu2023towards}; pseudo-labeling, in which the model's confident predictions serve as surrogate labels~\citep{chen2022contrastive}; and feature alignment, which matches test-time activation statistics to those of the source domain, as in ActMAD~\citep{mirza2023actmad}.

These recipes transfer poorly to MLLMs.
Backpropagation-based updates are computationally prohibitive at the scale of modern VLMs. Entropy- and pseudo-label-based objectives presuppose a fixed label space and calibrated uncertainty, whereas MLLMs generate open-ended text and remain confident when wrong~\citep{wang2025vision}. Online self-supervised updates carry a well-documented risk of model collapse and catastrophic forgetting~\citep{press2023rdumb, schirmer2025monitoring}.

Recent efforts extend TTA beyond classification to dense prediction and regression~\citep{adachi2025ssa, niu2025spa}. MLLMs, however, remain largely unaddressed. A large body of work exists for CLIP-like architectures~\citep{sheng2026illusion}, such as test-time prompt tuning~\citep{shu2022tpt}. These methods treat CLIP strictly as a zero-shot classifier, inheriting standard classification assumptions.

To the best of our knowledge, \citet{kaya2026efficient} is the only existing work on TTA for MLLMs. 
Their method aggregates predictions from augmented inputs at the token level and optionally updates parameters from the resulting pseudo-labels. However, it targets general test-time scaling for small MLLMs rather than robustness to input corruptions, and lacks evaluation under common corruptions. 
To address these gaps, we test whether activation engineering can provide a practical way to adapt MLLMs to corrupted images directly at test time.

\vspace{-4pt}
\paragraph{Activation Steering and Test-Time Feature Alignment}
Two largely independent lines of work modify a model's behavior at inference time by acting on its activations rather than its weights.
In the LLM community this is known as activation engineering or steering~\citep{zou2023representation}. 
The most notable Contrastive Activation Addition (CAA)~\citep{rimsky2024steering} computes a difference-of-means vector from contrastive prompt pairs and adds it to the residual stream.
In MLLMs, such interventions have so far been used to control high-level semantic behavior, e.g., to improve visual grounding or reduce hallucinations~\citep{liu2025reducing, khayatan2025analyzing, parekh2025learning}.
Recently Activation Transport (AcT)~\citep{rodriguez2025controlling, rodriguez2025lineas} reframes steering as optimal transport between source and target activation distributions.

In the TTA literature, the analogous operation---matching test-time features to those of the source domain---is called feature alignment~\citep{schneider2020improving}, and it underlies a recent wave of backpropagation-free methods that avoid the instabilities of gradient-based adaptation: FOA~\citep{niu2024foa} shifts the last-layer CLS token back toward source activations, NEO~\citep{murphy2026neo} re-centers test embeddings by a single vector, TCA~\citep{you2025tca} aligns correlations of cached representations, and PEA~\citep{xiao2026pea} realigns intermediate features via a layer-wise whitening--coloring transform (WCT).
Despite being lightweight and architecture-agnostic, all of these are developed and evaluated exclusively on image classifiers.

In this work we consider the Gaussian case of AcT (Gaussian-AcT), requiring only source statistics rather than source samples. It recovers existing feature-alignment methods as special cases---the re-centering of NEO and FOA without the scaling term and the WCT of PEA with full covariance.
To our knowledge, we are the first to repurpose activation steering for the low-level representation shifts induced by visual corruptions in MLLMs.

\vspace{-6pt}
\section{Experiments}
\label{sec:experiments}
\vspace{-4pt}
\vspace{-4pt}
\paragraph{Experimental Setup.}
We evaluate Qwen2.5-VL-3B-Instruct~\citep{bai2025qwen25vl}, InternVL3-2B~\citep{zhu2025internvl3}, and Ovis2-4B~\citep{lu2024ovis} on RealWorldQA~\citep{xai2025realworldqa}, a vision-centric multiple-choice question answering (MCQ) benchmark of 765 questions on real-world scenes, under the 15 common corruptions~\citep{hendrycks2019imagenet_c} at severity~5, with images resized to $896\times896$ pixels.
We use VLMEvalKit~\citep{duan2024vlmevalkit} to keep the evaluation protocol standardized and reproducible across all models and methods.
All three models share the LLaVA architecture~\citep{liu2023visual}: a vision encoder maps the image to patch embeddings, a projector maps each into the token-embedding space of a language model, and the language model processes the resulting \emph{visual tokens} together with the text.
Unlike a classifier, the language model reads every visual token rather than a single pooled vector; every method we study edits these token activations at test time and leaves all parameters frozen.
We edit vision encoder activations, where the corruption-induced shift originates and is still purely visual, so one correction estimated from images alone serves every question about the image.
The edited blocks, which we call the \emph{intervention layers}, and the remaining details are given in \Cref{app:experiments_details}.

\vspace{-4pt}
\paragraph{Gaussian-AcT.}
Let $\mu_{\mathrm{src}},\sigma_{\mathrm{src}}$ and $\mu_{\mathrm{tgt}},\sigma_{\mathrm{tgt}}$ be the per-dimension mean and standard deviation of the activations at an intervention layer on clean (source) and corrupted (target) images.
Under a Gaussian approximation with diagonal covariance, the optimal transport map from the corrupted to the clean distribution is the elementwise affine map
$T(\vx)=\mu_{\mathrm{src}}+(\sigma_{\mathrm{src}}/\sigma_{\mathrm{tgt}})\,(\vx-\mu_{\mathrm{tgt}})$,
the Gaussian case of activation transport (AcT)~\citep{rodriguez2025controlling}, which we call Gaussian-AcT.
The moments are estimated either \emph{globally}, one pair $(\mu,\sigma)$ per layer shared by all tokens, or \emph{per-token}, one pair per token position across images.
Both variants are \emph{sample-blind}: the same map is applied to every image and depends only on the token position, which is known at test time.

\label{sec:setup}

\vspace{-4pt}
\subsection{Online adaptation}
\label{sec:online}
\vspace{-3pt}

\begin{table}[t]\centering\small
\setlength{\tabcolsep}{2.5pt}
\vspace{-2pt}
\caption{\textbf{Online TTA with Gaussian-AcT barely moves accuracy on any of the three MLLMs.}
RealWorldQA accuracy (\%) under the 15 common corruptions at severity~5.
\emph{no-TTA} is the accuracy without adaptation.
The Gaussian-AcT columns give the change $\Delta$ over it in pp when applied at all intervention layers, without and with dynamic weighting.
Cells with $|\Delta|\ge2$~pp are shaded \swatch{teal!18}{green} for a gain and \swatch{red!15}{red} for a loss.
The better of the two corrections per corruption is in \textbf{bold}.
}
\label{tab:main_results}
\begin{tabular}{l ccc ccc ccc}
\toprule
& \multicolumn{3}{c}{InternVL3-2B} & \multicolumn{3}{c}{Qwen2.5-VL-3B} & \multicolumn{3}{c}{Ovis2-4B} \\
\cmidrule(lr){2-4}\cmidrule(lr){5-7}\cmidrule(lr){8-10}
& \hd{no-TTA} & \hd{\makecell{Gaussian-\\AcT}} & \hd{\makecell{+ dynamic\\weighting}} & \hd{no-TTA} & \hd{\makecell{Gaussian-\\AcT}} & \hd{\makecell{+ dynamic\\weighting}} & \hd{no-TTA} & \hd{\makecell{Gaussian-\\AcT}} & \hd{\makecell{+ dynamic\\weighting}} \\
\midrule
Clean & 62.5 & -- & -- & 65.6 & -- & -- & 69.4 & -- & -- \\
\midrule
\multicolumn{10}{l}{\emph{Noise}}\\
\quad Gaussian    & \nc{54.1}{0.4} & \dd{-0.5}{1.9} & \bd{+1.4}{0.8} & \nc{50.8}{0.5} & \dd{-4.1}{0.5} & \bd{+0.6}{1.0} & \nc{58.9}{0.3} & \dd{-1.7}{1.1} & \bd{-0.1}{0.5} \\
\quad Shot        & \nc{54.2}{0.5} & \dd{+0.4}{1.8} & \bd{+1.7}{1.1} & \nc{52.3}{0.1} & \dd{-3.7}{1.4} & \bd{-0.0}{0.9} & \nc{59.6}{0.6} & \bd{+1.0}{0.5} & \dd{+0.6}{0.5} \\
\quad Impulse     & \nc{55.0}{1.2} & \dd{+0.1}{1.3} & \bd{+0.7}{1.0} & \nc{52.2}{0.6} & \dd{-5.4}{1.0} & \bd{+0.8}{0.9} & \nc{59.5}{1.9} & \bd{+0.4}{2.0} & \dd{+0.4}{1.7} \\
\addlinespace[2pt]
\multicolumn{10}{l}{\emph{Blur}}\\
\quad Defocus     & \nc{55.3}{0.0} & \dd{+1.2}{0.3} & \bd{+2.7}{0.2} & \nc{53.9}{0.0} & \dd{-0.1}{1.3} & \bd{+3.4}{0.8} & \nc{59.0}{0.0} & \bd{+4.0}{0.1} & \dd{+3.4}{0.1} \\
\quad Glass       & \nc{58.0}{0.3} & \bd{+0.6}{0.6} & \dd{+0.3}{0.5} & \nc{57.7}{0.5} & \dd{-2.1}{1.9} & \bd{+0.7}{0.5} & \nc{61.1}{0.3} & \bd{+1.9}{0.8} & \dd{+1.4}{0.7} \\
\quad Motion      & \nc{46.1}{0.9} & \bd{+1.0}{0.6} & \dd{+0.7}{0.5} & \nc{46.7}{0.6} & \dd{-0.7}{1.1} & \bd{+0.9}{1.4} & \nc{50.5}{1.1} & \dd{+0.7}{2.1} & \bd{+0.8}{2.1} \\
\quad Zoom        & \nc{48.6}{0.0} & \bd{+3.6}{0.3} & \dd{-0.1}{0.9} & \nc{46.4}{0.0} & \bd{+3.6}{0.4} & \dd{+1.7}{0.7} & \nc{51.6}{0.0} & \bd{+6.1}{0.2} & \dd{+3.8}{0.6} \\
\addlinespace[2pt]
\multicolumn{10}{l}{\emph{Weather}}\\
\quad Snow        & \nc{53.4}{0.7} & \dd{+0.5}{0.4} & \bd{+1.3}{0.5} & \nc{51.8}{0.6} & \dd{-0.5}{1.1} & \bd{+1.0}{0.9} & \nc{56.0}{0.3} & \bd{+1.0}{0.6} & \dd{+0.8}{0.5} \\
\quad Frost       & \nc{56.5}{1.9} & \dd{-0.3}{2.2} & \bd{+0.9}{0.7} & \nc{55.8}{0.1} & \dd{-1.4}{1.1} & \bd{+0.8}{0.2} & \nc{61.0}{0.5} & \dd{+0.7}{0.5} & \bd{+1.0}{0.2} \\
\quad Fog         & \nc{59.0}{0.7} & \bd{+1.4}{0.2} & \dd{+1.1}{0.4} & \nc{57.1}{1.2} & \dd{+1.9}{2.2} & \bd{+4.0}{2.7} & \nc{63.7}{0.2} & \bd{+2.6}{0.9} & \dd{+2.6}{0.7} \\
\quad Brightness  & \nc{60.9}{0.0} & \bd{-1.0}{0.3} & \dd{-1.2}{0.3} & \nc{61.6}{0.0} & \dd{-4.0}{0.5} & \bd{-1.6}{0.2} & \nc{68.5}{0.0} & \dd{-1.5}{0.4} & \bd{-1.2}{0.8} \\
\addlinespace[2pt]
\multicolumn{10}{l}{\emph{Digital}}\\
\quad Contrast    & \nc{52.3}{0.0} & \bd{+7.1}{0.7} & \dd{+5.9}{0.6} & \nc{50.1}{0.0} & \dd{+0.7}{0.7} & \bd{+3.1}{0.2} & \nc{57.0}{0.0} & \bd{+9.4}{1.4} & \dd{+8.1}{0.1} \\
\quad Elastic     & \nc{56.6}{0.2} & \dd{+0.4}{0.9} & \bd{+0.9}{1.0} & \nc{54.7}{0.6} & \dd{-0.3}{1.2} & \bd{+2.5}{0.9} & \nc{61.1}{0.5} & \bd{+3.0}{1.1} & \dd{+2.9}{0.7} \\
\quad Pixelate    & \nc{46.5}{0.0} & \dd{-1.3}{0.3} & \bd{-0.2}{0.3} & \nc{49.0}{0.0} & \dd{-4.5}{0.8} & \bd{-0.4}{0.5} & \nc{52.3}{0.0} & \dd{-0.7}{1.0} & \bd{+0.9}{0.3} \\
\quad JPEG        & \nc{59.5}{0.0} & \bd{-0.7}{0.3} & \dd{-1.0}{1.0} & \nc{59.3}{0.0} & \dd{-5.0}{0.9} & \bd{-1.4}{1.1} & \nc{65.5}{0.0} & \dd{-2.5}{0.2} & \bd{-0.9}{0.4} \\
\midrule
\textbf{Mean} & \nc{54.4}{0.1} & \dd{+0.8}{0.3} & \bd{+1.0}{0.3} & \nc{53.3}{0.1} & \dd{-1.7}{0.4} & \bd{+1.1}{0.3} & \nc{59.0}{0.2} & \bd{+1.6}{0.3} & \dd{+1.6}{0.4} \\
\bottomrule
\end{tabular}
\vspace{-8pt}
\end{table}

In online adaptation, the target statistics of the Gaussian-AcT map are estimated from the test stream itself, given only the precomputed source statistics, so the correction should improve as more corrupted data arrives.
Each corruption forms a separate stream and the target statistics are reset between corruptions.
We report means over three stream orders with the between-order standard deviation.
We apply the map with per-token moments at every intervention layer, either unweighted or with \emph{dynamic weighting}, adapted from PEA~\citep{xiao2026pea}, which scales the correction of each layer by how far its running target statistics have drifted from the source statistics (more details~in~\Cref{app:online}).

\vspace{-4pt}
\paragraph{Unweighted correction is unreliable, weighted correction marginal.}
Averaged over the 15 corruptions (\Cref{tab:main_results}), the unweighted correction gains $+0.8$ and $+1.6$~pp on InternVL3-2B and Ovis2-4B but loses $1.7$~pp on Qwen2.5-VL-3B, and individual corruptions swing by several points in either direction.
Dynamic weighting lifts Qwen2.5-VL-3B to $+1.1$~pp and leaves the other two models unchanged at $+1.0$ and $+1.6$~pp.
These gains recover $12\%$, $9\%$, and $15\%$ of the $8.1$, $12.3$, and $10.4$~pp that corruption costs each model.
The weighting achieves this by switching the correction off rather than by improving it: the weight concentrates on the earliest layers, and accuracy improves only gradually as the stream progresses (\Cref{fig:dynamic_weight,fig:cumulative}).

\vspace{-4pt}
\paragraph{What little is recovered comes from globally acting corruptions.}
Contrast ($+5.7$~pp averaged over models), defocus blur ($+3.2$), fog ($+2.6$), and zoom blur ($+1.8$) account for most of the gain, whereas JPEG compression ($-1.1$) and pixelate ($+0.1$) are no better than no-TTA on any model.
All four corruptions act smoothly over the whole image, either as a photometric change (contrast, fog) or as a low-frequency blur (defocus, zoom), while contrast is the only one whose gap is more than half recovered, on two of the three models.
High-frequency, spatially local corruptions are left almost uncorrected.
Brightness is the apparent exception, losing $1.3$~pp, but at severity~5 it costs these models only $0.9$--$4.0$~pp in the first place, leaving nothing to recover.
\Cref{sec:analysis} traces this split to the structure of the shift itself.

Online adaptation thus yields no gain of practical significance: unweighted it can hurt, and weighted it barely helps.
The feature alignment that works for image classification does not carry over to MLLMs, even when they adapt to one corruption at a time.

\subsection{The method is not the bottleneck}
\label{sec:upper_bound}

\begin{figure}[t]
  \centering
  \includegraphics[width=\linewidth]{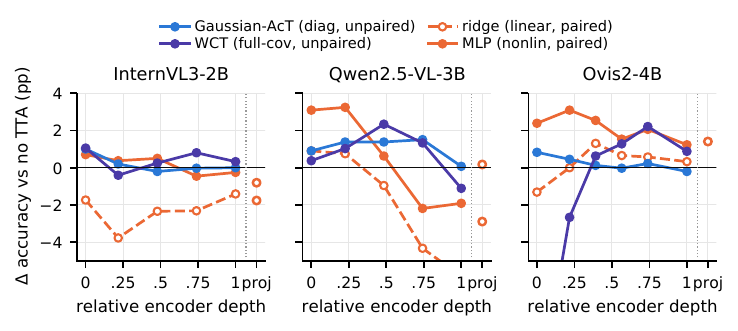}
  \vspace{-4pt}
  \caption{\textbf{A more expressive map does not recover more.} Change in accuracy over the no-TTA baseline, averaged over the 15 corruptions, when a single layer is corrected. Gaussian-AcT and WCT are fitted from distribution statistics. Ridge and MLP on paired clean--corrupted tokens.}
  \label{fig:upper_bound}
\vspace{-8pt}
\end{figure}

Gaussian-AcT is deliberately simple: a diagonal affine map fitted to distribution statistics estimated online.
A null result could therefore be attributed to the estimator, to the expressivity of the map, to its supervision, or to the site of the intervention.
We rule out each in turn.

\vspace{-4pt}
\paragraph{Online estimation is not the bottleneck.}
At the single intervention block, per-token moments estimated online match the same map fitted offline on the full test set, with per-corruption gains correlated at $r=0.93$ (\Cref{fig:online_vs_offline}), and pooling the moments over token positions lowers the gain on every model (\Cref{fig:online_cumulative}).
Per-token moments are therefore preferable, yet they still recover only about~one~point.

\vspace{-4pt}
\paragraph{Expressivity and supervision are not the bottleneck.}
We next drop the online constraint and fit maps offline, with clean and corrupted images available together, and vary \textbf{expressivity} (linear vs.\ nonlinear) and \textbf{supervision} (unpaired statistics vs.\ \emph{paired} tokens, each corrupted token regressed onto its own clean counterpart).
Paired supervision is unavailable to any test-time method, so these results bound what such a method could achieve rather than propose one.
Every map is shared across token positions and takes only the token's own activation as input.
Gaussian-AcT and its full-covariance counterpart WCT are fitted from distribution statistics, whereas a closed-form ridge regression and a two-layer MLP are fitted on paired tokens of 500 images.
All maps are evaluated on the remaining 265 held-out images.

\Cref{fig:upper_bound} shows that neither factor changes the outcome.
The oracle MLP recovers at most a quarter of the held-out gap ($+0.50$, $+3.25$, and $+2.54$~pp against gaps of $5.8$, $15.2$, and $10.7$~pp), and only at a shallow layer.
WCT stays within one standard error of Gaussian-AcT despite fitting a full covariance, and the paired ridge map lowers accuracy at most layers despite its oracle supervision: least squares returns the conditional mean of the clean token given the corrupted one, so the detail it cannot predict is shrunk away.
What a single-layer, position-shared correction can recover is thus bounded regardless of estimator, expressivity, or supervision.

The MLP's failure says more than this bound.
Its correction is a nonlinear function of the input token, but it uses neither the token's position, which per-token moments already capture, nor information from the other tokens of the image.
The shift of a token is therefore not recoverable from that token alone. Undoing it requires the information from the rest of the image, which places the problem closer to restoration than to feature alignment.

\vspace{-4pt}
\paragraph{The site is not the bottleneck.}
At the intervention layers, interpolating each token's activation from its corrupted toward its clean value recovers accuracy in proportion to the interpolation weight (\Cref{app:interpolation}), so a perfect per-token correction at a single layer would close the gap.
Nor does the damage concentrate elsewhere in the encoder: activation patching, which overwrites the output of one encoder sub-module, the attention or MLP branch of a block, with its clean-image activation, upper-bounds what any adaptation applied there can achieve, and except for Ovis2's projector-side MLP no single sub-module recovers more than about half of the gap, at a location that depends on the architecture (\Cref{app:activation_patching}).
What sample-blind adaptation lacks is therefore not a place to intervene but a correction specific to each token.

\vspace{-6pt}
\section{Analysis}
\label{sec:analysis}
\vspace{-4pt}
\subsection{The shift is token-specific}
\label{sec:shift_analysis}

\begin{figure}[t]
  \centering
  \includegraphics[width=\linewidth]{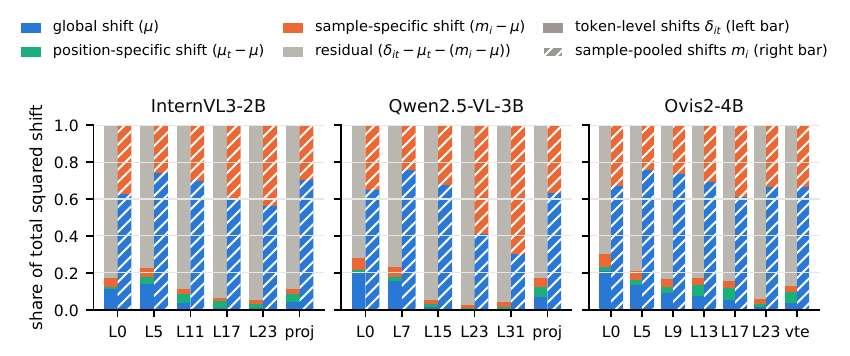}
  \vspace{-4pt}
  \caption{\textbf{Decomposition of the total squared shift (\Cref{eq:shift_decomposition})
  across layers}, averaged over the 15 corruptions. Solid bars: token-level shifts; hatched bars:
  sample-pooled shifts. The residual dominates the token-level shift at every layer,
  whereas the global shift dominates the pooled shift almost everywhere.}
  \label{fig:decomposition}
\vspace{-8pt}
\end{figure}

The limited gains above raise a simpler question: how much of the corruption-induced shift could a correction of a given granularity remove at all, whatever its estimator?
Let $\delta_{it}$ be the shift of token $t$ of sample $i$ between its corrupted and clean activation ($N$ samples, $T$ tokens each), and let $\mu$, $\mu_t$, and $m_i$ be its means over all tokens, over the samples at position $t$, and over the tokens of sample $i$.
A correction can remove at most the part of the shift that is constant within its granularity: a global correction removes $\mu$, a per-token correction also removes the position-specific part $\mu_t-\mu$, and a per-sample correction, additionally removes the sample-specific part $m_i-\mu$.
What remains varies from token to token within an image.
The four parts are orthogonal under summation over $i$ and $t$, because each centered part sums to zero over the index it is centered on, so they partition the total squared shift exactly:
\begin{equation}
\label{eq:shift_decomposition}
  \sum_{i,t}\normtwo{\delta_{it}}^{2}
  =
  \underbrace{NT\normtwo{\mu}^{2}}_{\text{global}}
  + \underbrace{N\sum_t\normtwo{\mu_t-\mu}^{2}}_{\text{position-specific}}
  + \underbrace{T\sum_i\normtwo{m_i-\mu}^{2}}_{\text{sample-specific}}
  + \underbrace{\sum_{i,t}\normtwo{\delta_{it}-\mu_t-m_i+\mu}^{2}}_{\text{residual}}.
\end{equation}
Each share is the largest fraction of the total squared shift that the corresponding correction can remove.
It bounds the shift removed, not the accuracy recovered, which \Cref{sec:upper_bound} measures directly.

\Cref{fig:decomposition} reports the four shares at every intervention layer.
The global share is modest at the first blocks and vanishes with depth; conditioning on token position adds little, and even a per-sample oracle recovers only a small additional fraction.
The residual, the variation from token to token within the same image, accounts for the large majority of the shift everywhere.
Corruption thus changes the content of individual tokens rather than displacing their distribution as a whole, which explains why sample-blind corrections help only at the shallowest blocks and only for globally acting corruptions (\Cref{app:per_corruption}).

\begin{figure}[t]
  \centering
  \includegraphics[width=\linewidth]{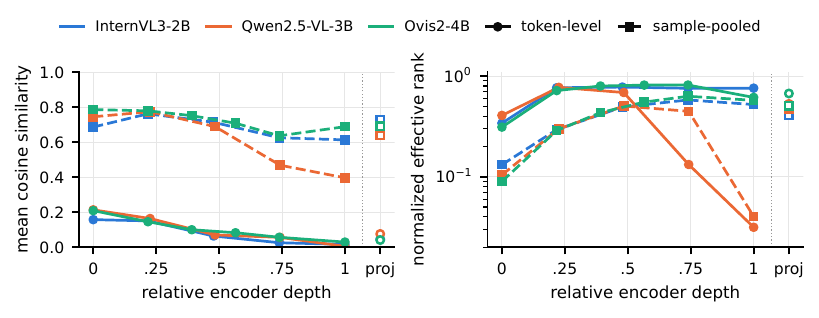}
  \vspace{-4pt}
  \caption{\textbf{Token-level shifts (solid) and sample-pooled shifts (dashed) versus encoder
  depth.} \textbf{(a)}~Mean cosine similarity between the shifts of different images and
  \textbf{(b)}~normalized effective rank~\citep{roy2007effective} of the shift matrix, averaged over
  the 15 corruptions.}
  \label{fig:cos_rank}
\vspace{-8pt}
\end{figure}

\Cref{fig:cos_rank} confirms that there is no shared direction to exploit.
The token-level shifts of different images are only weakly aligned at the first block and become nearly orthogonal with depth. 
Also, the matrix of token shifts keeps a high normalized effective rank~\citep{roy2007effective} throughout the encoder.
The shifts of a corruption thus collectively span almost the full embedding space, leaving no low-dimensional structure for a shared correction to exploit.

\vspace{-4pt}
\paragraph{Why the same methods work for classification.}
Feature alignment succeeds in image classification, where the prediction is read from a single pooled representation per image, the CLS token or mean-pooled activations, whether the correction is applied to that vector alone~\citep{niu2024foa,murphy2026neo} or to every token of the intermediate layers~\citep{xiao2026pea}.
Averaging the token-level shifts of each image reproduces this view: the position-specific and residual components average out, and \Cref{eq:shift_decomposition} reduces to $\sum_i\normtwo{m_i}^{2} = N\normtwo{\mu}^{2}+\sum_i\normtwo{m_i-\mu}^{2}$.
In this pooled view the decomposition inverts (hatched bars in \Cref{fig:decomposition}, dashed lines in \Cref{fig:cos_rank}).
Pooled shifts of different images are strongly aligned and span few directions, and the global shift $\mu$ accounts for the majority of the squared pooled shift at almost every intervention layer of all three models, so a single sample-blind shift captures most of what there is to correct.
The residual that dominates the token-level shift is thus not removed by pooling, only averaged out of the vector a classifier reads, while the language model reads the tokens themselves, residual included.

The pooled view is also the one in which the correction works (\Cref{fig:teaser}; details and per-corruption results in \Cref{app:probe}).
A linear probe trained on clean, mean-pooled projector activations to recognize five scene types falls from $94.5\%$ to $57.6\%$ accuracy under corruption, averaged over models.
Gaussian-AcT fitted to the pooled activations lifts it to $84.0\%$, recovering a fraction $\rho=0.65$--$0.78$ of the probe's clean--corrupted gap and improving every model--corruption pair.
A map of the same form, fitted on the same corrupted images at the same projector output but applied to every visual token, recovers none of the RealWorldQA accuracy gap ($\rho$ between $-0.04$ and $0.04$).
The corrupted representation still contains the scene information, and matching its first two moments makes that information linearly decodable again.
A classifier that reads a pooled feature benefits from this; an MLLM whose language model attends to every visual token does not.

\vspace{-4pt}
\subsection{The degradation is not localized}
\label{sec:substitution}
\vspace{-3pt}

\begin{wrapfigure}{r}{0.4\linewidth}
    \vspace{-10pt}
    \centering
    \includegraphics[width=\linewidth]{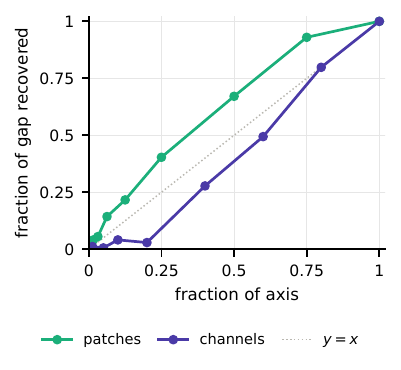}
    \vspace{-4pt}
    \caption{\textbf{No small subset of the representation carries the degradation.} The fraction of the clean--corrupted gap
    recovered on both axes (dotted: proportional), for Qwen2.5-VL-3B under Gaussian noise at
    severity~5.}
\label{fig:substitute_neurons_patches_qwen}
    \vspace{-10pt}
\end{wrapfigure}

Prior work shows that the task-relevant visual signal in MLLMs is highly concentrated: most visual
tokens can be pruned with little accuracy loss~\citep{chen2024image_is_worth}, and cropping to a single question-relevant region recovers much of the accuracy lost on fine-grained VQA~\citep{zhang2024perceivingsmallvisualdetails,wu2024guided_visual_search}. 
If the degradation caused by corruption were similarly concentrated, repairing a small subset of the representation would suffice, and correction would reduce to a selection problem.

We test this with an oracle intervention at the final encoder block of all three models, under Gaussian noise at severity~5. 
We replace a random subset of the corrupted activation with the clean activation of the same image and sweep the size of the subset along two axes: whole token positions, and feature dimensions across all positions. 

Recovery requires restoring a large fraction of the representation on either axis (\Cref{fig:substitute_neurons_patches_qwen}): half of the gap costs roughly a third of all patches or
60\% of all feature dimensions. 
For Qwen2.5-VL-3B, the two axes differ in shape, while for the remaining two models the curves nearly align (\Cref{fig:substitute_neurons_patches_intern3,fig:substitute_neurons_patches_ovis2}) having a steeper slope at lower fractions. 
Patch recovery is concave and begins immediately, whereas dimension recovery is flat below about 20\% of the dimensions and stays below the proportional rate throughout: corruption is spread across nearly all feature dimensions and somewhat unevenly across positions. 
Choosing the subset by the magnitude or the variance of the clean activations instead of at random rescales these curves by at most a constant factor and leaves their shape unchanged (\Cref{app:selection}). 
The concentration that token pruning and visual cropping exploit therefore does not extend to the degradation caused by corruption: no small region of the representation carries it, and no selection rule we tested finds one.

\vspace{-6pt}
\section{Discussion, Limitations, and Future Work}
\vspace{-4pt}

Across three MLLMs and 15 corruptions, representation-based TTA recovers about one percentage point of a drop an order of magnitude larger, and no variant of the method changes this: online or offline estimation, global or per-token moments, linear or nonlinear maps, and unpaired or paired supervision recover at most a few points (\Cref{sec:online,sec:upper_bound}).
Prompting about the degradation does not help either, and the confidence signal on which entropy-based objectives rely barely predicts correctness (\Cref{app:prompting,app:entropy}).
The reason is the structure of the shift, not the method: it is almost entirely token-specific (\Cref{sec:shift_analysis}), so a sample-blind correction has little to correct, whereas pooled per image, as a classifier reads it, the same shift is dominated by a shared direction, which is why feature alignment works for classification.
The correction that would work must be specific to each sample and token, and no small subset of tokens or dimensions carries it (\Cref{sec:substitution}).
Adapting MLLMs to corruptions is therefore a per-token restoration problem rather than a feature-alignment problem, and sample-blind corrections should not be expected to repair corruption damage at deployment.

\vspace{-4pt}
\paragraph{Limitations.}
We test only sample-blind corrections: none is estimated from the image it corrects, and none draws on the rest of the image.
The study is preliminary in scale: three models of 2--4B parameters whose language models come from one family (Qwen2.5), and one multiple-choice benchmark of 765 questions.
Corrections are applied in the vision encoder and projector only. The corruptions are synthetic, at a single severity, and one per stream, rather than the mixed degradations~met in deployment.

\vspace{-4pt}
\paragraph{Future work.}
MAA trains layer-wise adapters in a frozen vision encoder that steer corrupted features back onto the clean manifold, using paired clean--corrupted features and a token-level adversarial constraint~\citep{li2026robust}.
A test-time counterpart would have to learn such a token-level correction from the unlabeled stream alone, for instance through consistency across augmentations~\citep{kaya2026efficient}.
Whether this is possible without paired data is the question we leave open.

\begin{ack}
We gratefully acknowledge Polish high-performance computing infrastructure PLGrid (HPC Center: ACK Cyfronet AGH) for providing computer facilities and support within computational grant no. PLG/2025/018644 and PLG/2026/019363.
This research was funded in whole or in part by National Science Centre, Poland, grant no \mbox{2023/51/D/ST6/02846} and \mbox{2024/53/N/ST6/03156}. Bartłomiej~Twardowski acknowledges the grant RYC2021-032765-I.
\end{ack}

\bibliographystyle{plainnat}
\bibliography{bibliography}

\appendix
\raggedbottom
\section{Experimental Details}
\label{app:experiments_details}

\paragraph{Intervention layers.}
Unless stated otherwise, we intervene at the outputs of encoder blocks $\{0,5,11,17,23\}$ and the projector of InternVL3-2B, blocks $\{0,7,15,23,31\}$ and the projector of Qwen2.5-VL-3B, and blocks $\{0,5,9,13,17,23\}$ and the visual embedding table of Ovis2-4B, which span each encoder at roughly equal relative depth; we refer to these as the \emph{intervention layers}.
Qwen2.5-VL and InternVL3 merge neighboring patch embeddings before projection, and Ovis2 replaces the projector with a lookup in a learned visual embedding table (\emph{vte}).
Single-layer experiments use blocks 11, 11, and 9, respectively, selected by activation patching (\Cref{app:activation_patching}).
Source statistics are computed once on clean images, while target statistics are estimated either offline on the full test set or online from a test stream (\Cref{app:online}).
We report accuracy averaged over three random seeds, which for online adaptation also determine the sample order.

\paragraph{Source statistics.}
\label{app:source_stats}
Our use of pre-computed source statistics ($\mu_{\mathrm{src}},\sigma_{\mathrm{src}}$) sits at the boundary between test-time adaptation and unsupervised domain adaptation and deserves comment. In the standard TTA setting, the model pre-trained on the source domain is available at test time but the source data are not~\citep{liang2025comprehensive}. Storing summary statistics of the source activations is a one-time offline computation: it requires a single forward pass over a calibration set and a few tensors per layer, on the order of tens of megabytes for a ViT-scale encoder, and grants no access to individual source samples at adaptation time. The same convention is followed by prior feature-alignment methods: ActMAD~\citep{mirza2023actmad} aligns test activations with source activation statistics, FOA~\citep{niu2024foa} measures the discrepancy between test and source statistics and shifts activations to reduce it, and PEA~\citep{xiao2026pea} pre-computes per-block source statistics offline. We adopt it throughout. Retaining these statistics is a mild relaxation of settings that prohibit retaining any information derived from the source domain; the adaptation procedure nevertheless never sees a source example and incurs only a small, fixed storage cost.

\section{Online Estimation and Dynamic Weighting}
\label{app:online}

\begin{figure}[t]
  \centering
  \includegraphics[width=\linewidth]{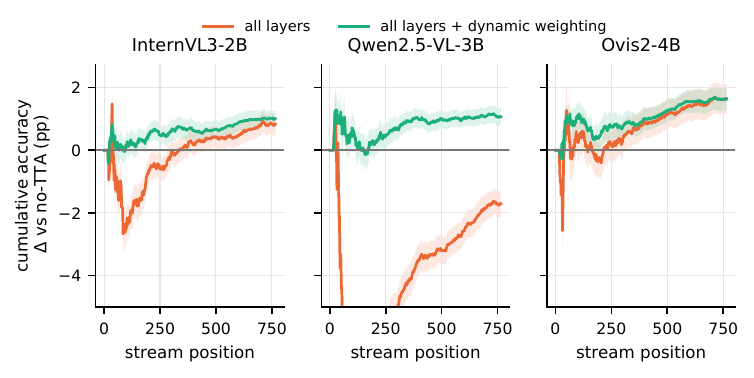}
  \caption{\textbf{Adaptation is slow and the gain stays small.} Cumulative change in RealWorldQA
  accuracy over the no-TTA baseline along the stream (mean over corruptions and three orders; band:
  standard error). The unweighted correction drops sharply early, while the online estimate is
  still poor, and recovers to about break-even; the weighted variant stays near zero and ends with
  a small gain.}
  \label{fig:cumulative}
\end{figure}

\begin{figure}[t]
  \centering
  \includegraphics[width=0.45\linewidth]{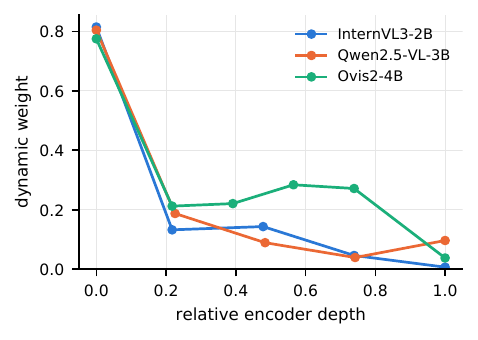}
  \caption{\textbf{The weight concentrates on the early layers.} Mean per-layer weight over the
  stream against relative encoder depth.}
  \label{fig:dynamic_weight}
\end{figure}

Target moments are estimated online in batches of $B = 16$ samples: after each batch, the running mean and variance of every intervention layer are updated by an exponential moving average with rate $0.1$.
\Cref{fig:cumulative} traces the cumulative accuracy gain along the stream and
\Cref{fig:dynamic_weight} the per-layer weights of the dynamically weighted variant of
\Cref{sec:online}.

\paragraph{Dynamic weighting.}
The weight of layer $\ell$ follows the per-layer rule of PEA~\citep{xiao2026pea}: the discrepancy
$d_\ell$, the Kullback--Leibler divergence between the Gaussian fits of the running target moments
and of the source moments, is rescaled across layers to $[0,1]$ by min--max normalization, so that the most discrepant layer is corrected at full strength
and the least discrepant one not at all. Because deeper layers begin estimating their moments later
in our online cascade, a layer with a nearly empty estimate would otherwise appear maximally
discrepant and receive full weight. We therefore multiply the weight by a confidence factor
$n_\ell/(n_\ell+B)$, where $n_\ell$ is the number of samples the estimate of layer $\ell$ has
seen, which keeps the weight near zero until the estimate rests on data.

\section{Ablations of Single-Layer Gaussian-AcT}
\label{app:single_layer_ablation}

We run both ablations on single-layer Gaussian-AcT fitted online (\Cref{sec:online}) at
block 11 of InternVL3-2B and Qwen2.5-VL-3B and block 9 of Ovis2-4B, under the 15 corruptions at
severity~5, and report the whole-episode accuracy gain over the no-TTA baseline as mean and standard
deviation over three stream orders.

\begin{figure}[t]
  \centering
  \includegraphics[width=0.5\linewidth]{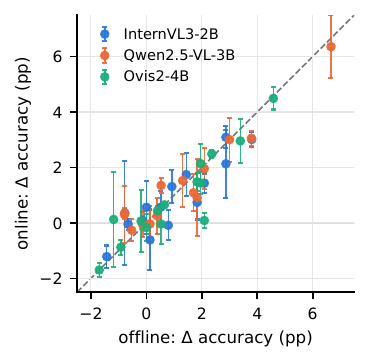}
  \caption{\textbf{Online estimation matches offline fitting.} Whole-episode accuracy gain over
  the no-TTA baseline of per-token Gaussian-AcT with target moments fitted offline ($x$) or
  estimated online ($y$); one point per model and corruption (mean over three stream orders; bars:
  standard deviation); dashed line: identity.}
  \label{fig:online_vs_offline}
\end{figure}

\paragraph{Online fitting matches offline fitting.}
With per-token moments estimated online, accuracy improves by $+0.87\pm0.38$, $+1.32\pm0.23$, and
$+0.91\pm0.10$~pp on InternVL3-2B, Qwen2.5-VL-3B, and Ovis2-4B, compared with $+0.95$, $+1.27$, and
$+1.05$~pp when the moments are fitted offline on the full test set. Per corruption
(\Cref{fig:online_vs_offline}), the online and offline gains are strongly correlated ($r=0.93$) and
differ by $-0.06$~pp on average (SD~$0.64$).

\begin{figure}[t]
  \centering
  \includegraphics[width=\linewidth]{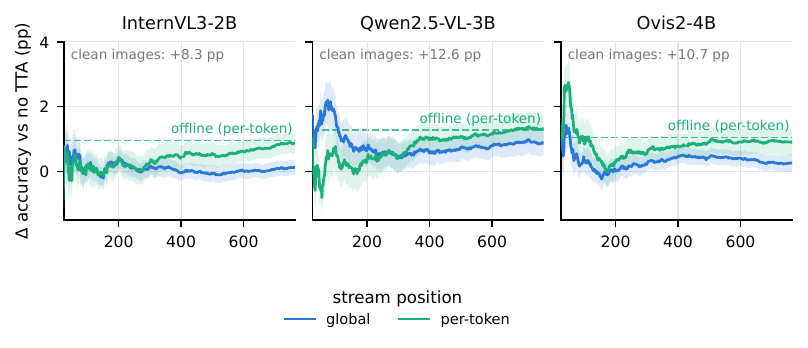}
  \caption{\textbf{Per-token moments converge more slowly but higher.} Cumulative accuracy gain
  over the no-TTA baseline along the stream for single-layer Gaussian-AcT fitted online with
  global and per-token target moments, averaged over the 15 corruptions (band: $\pm1$ standard
  error over corruptions; each corruption averaged over three stream orders). The dashed line marks
  the same per-token Gaussian-AcT fitted offline and the inset text the gain of clean images; the
  first 16 samples are not adapted.}
  \label{fig:online_cumulative}
\end{figure}

\paragraph{Per-token moments outperform global moments.}
Pooling the moments over all token positions reduces the gain to $+0.12\pm0.17$, $+0.88\pm0.14$,
and $+0.25\pm0.08$~pp. \Cref{fig:online_cumulative} shows the gain along the stream: the global
estimate converges within the first 50 samples but plateaus, whereas the per-token estimate
converges after 200--300 samples and reaches the accuracy of offline fitting. For reference, the
corruptions lower accuracy by 8.3, 12.6, and 10.7~pp relative to clean images.

\section{Activation Patching: Where the Degradation Can Be Repaired}
\label{app:activation_patching}

Activation patching~\citep{heimersheim2024patching} overwrites activations during a forward pass
with activations cached from a pass on a different input and observes how the output changes. Here
the pass is on a corrupted image, the cached activations come from the same image before
corruption, and we overwrite the output of one encoder sub-module, the attention or MLP branch of
one block, at a time. We measure the share of the lost answer margin, the logit of the correct
option minus the mean logit of the incorrect ones, that returns. Because the clean activation is the
best possible correction for that sub-module, this share upper-bounds what any adaptation confined
to it can achieve, whatever method estimates the correction.

Concretely, for each corrupted image we run the model once per sub-module $u$ with a forward hook
that replaces its output for all tokens by the cached clean activation, and record the answer
margin $M_u = \ell_{\mathrm{correct}} - \operatorname{mean}_{k}\ell_{k}$ over the option-letter
logits at the answer position, with $k$ ranging over the incorrect options. We report
$(M_u - M^{\mathrm{corr}})/(M^{\mathrm{clean}} - M^{\mathrm{corr}})$ averaged over 100 images per
corruption for all 15 corruptions at severity~5. Normalizing by the gap makes cells comparable
across models and corruptions, at the price of noisier rows for corruptions with a small gap
(brightness). Replacing the output of a whole block, which cuts the entire visual stream,
reproduces the clean margin exactly in every run and validates the setup.

\begin{figure}[t]
  \centering
  \includegraphics[width=0.85\linewidth]{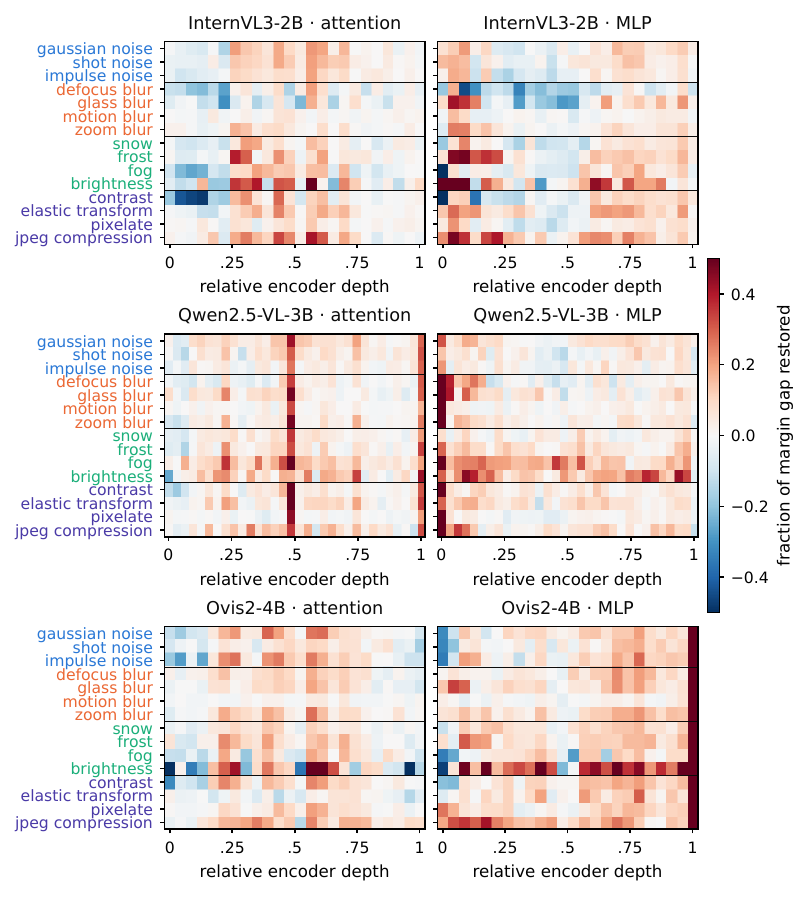}
  \caption{\textbf{Per-corruption encoder repair map.} Share of the
  clean--corrupted margin gap recovered by patching a single attention (left) or
  MLP (right) sub-module to its clean activation, for all 15 corruptions at
  severity~5 (rows, colored by family) and every encoder block (columns,
  relative depth); 100 images per cell, color saturates at $\pm0.5$.}
  \label{fig:patching_all}
\end{figure}

\Cref{fig:patching_all} shows that repair is localized but that its location depends on the
architecture. Qwen2.5-VL-3B recovers most low-frequency degradation (blur, digital) in its \emph{first}
MLP and has two corruption-general attention hubs, at mid-depth and at the end of the encoder.
Ovis2-4B's repair is broad rather than concentrated: a mid-depth attention cluster and the MLPs of the
final blocks are positive for nearly all corruptions. Its projector-side MLP, the MLP of its last block, appears to restore almost the
entire gap, but it accounts for most of the norm of its block's output, so the effect cannot be
attributed to the MLP rather than to the restoration of the visual stream as a whole. InternVL3-2B has
no comparable lever: recoverable degradation is spread thinly over mid-depth blocks, and restoring
early sub-modules in isolation is harmful for low-frequency corruptions, so their effect is not
captured by any single sub-module. Except for the projector-side MLP of Ovis2-4B, no single sub-module
recovers more than about half of the gap.

The single-layer experiments (\Cref{sec:online} and \Cref{app:single_layer_ablation}) adapt block 11 of InternVL3-2B and Qwen2.5-VL-3B and block 9 of Ovis2-4B, the upstream edge of the mid-depth region whose attention sub-modules recover damage for nearly every corruption (\Cref{fig:patching_all}).
Intervening there places the correction where repair is corruption-general and upstream of the encoder's own repair machinery, rather than at the strongest single component, which is corruption-specific in Qwen2.5-VL-3B and sits at the very end of the encoder in Ovis2-4B.
The maps also set expectations: a single-block correction has more headroom where repair is concentrated (Qwen2.5-VL-3B, Ovis2-4B) than in InternVL3-2B, where it is spread thinly.

\section{Interpolating Between Corrupted and Clean Activations}
\label{app:interpolation}

We evaluate Qwen2.5-VL-3B on 200 random samples, for each of which both the clean
(source) and the corrupted (target) image are available. For every sample we linearly interpolate,
at a given site, between the activation of the corrupted image and that of the clean image,
\begin{equation}
  y(\alpha) = (1-\alpha)\,\vx^{\mathrm{target}} + \alpha\,\vx^{\mathrm{source}},
\end{equation}
sweep $\alpha\in[0,1]$, and measure downstream accuracy. $\alpha=0$ reproduces the no-TTA baseline
and $\alpha=1$ the clean baseline; the sweep therefore traces how accuracy depends on the precision
of the correction at that site, independently of any estimator's error.
We apply it at the visual tokens at the output of an encoder block or of the projector, sweeping the site over blocks 4, 18, and 30 and the projector, under Gaussian noise at severity~5.

\begin{figure}[t]
  \centering
  \includegraphics[width=0.45\linewidth]{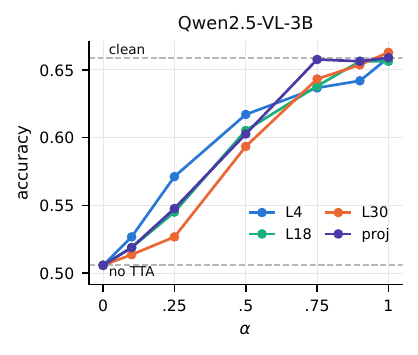}
  \caption{\textbf{A perfect single-layer correction recovers clean accuracy.} Accuracy of
  Qwen2.5-VL-3B under Gaussian noise at severity~5 when the activation of one site is
  interpolated between its corrupted ($\alpha=0$) and clean ($\alpha=1$) value; one curve per
  layer, dashed line: clean accuracy.}
  \label{fig:interpolation}
\end{figure}

\paragraph{Does the interpolation leave the manifold?}
\citet{li2026robust} argue that aligning corrupted features with clean references in Euclidean space is not semantics-preserving: the features move geometrically closer to the clean ones while drifting off the in-distribution support on which the language model's reasoning is calibrated, and this motivates the adversarial manifold constraint of their Manifold-Adversarial Adapters (MAA).
The interpolation above is such a Euclidean operation.
Every intermediate point is a convex combination of the corrupted activation and the clean activation of the same token, and nothing constrains it to lie on either the clean or the corrupted manifold.
If the segment left the support the language model handles, accuracy should dip or stall at intermediate $\alpha$.
\Cref{fig:interpolation} shows no such dip.
Accuracy grows roughly in proportion to $\alpha$ at every site, no intermediate $\alpha$ falls below the no-TTA baseline, and the curves meet the clean accuracy at $\alpha=1$.
This suggests that the straight path from a corrupted token to its own clean counterpart stays within the region the language model reads correctly, at least for this model and corruption.
We do not see this as contradicting MAA.
The drift they describe arises when a feature is pulled toward a clean reference that is not its paired counterpart, as in distribution-level alignment, where the target is a mean or a nearby clean feature rather than the sample's own clean activation.
Read this way, the obstacle is the sample-blind target rather than the Euclidean geometry of the correction, which is consistent with \Cref{sec:shift_analysis}.
Two caveats apply.
We do not measure manifold membership directly but infer it from the accuracy along the path, and the sweep covers one model and one corruption.

\section{Linear Probe on Pooled Projector Activations}
\label{app:probe}

This section details the probe experiment summarized in \Cref{sec:shift_analysis}. Under
corruption, answer accuracy confounds two failures: the visual features may lose the scene
information, or the language model may fail to use it. We isolate the first by scoring the
projector output directly with a linear probe. RealWorldQA~\citep{xai2025realworldqa} carries no
scene labels, so we construct them. We mean-pool the projector output over all visual tokens
($D{=}2048$ for Qwen2.5-VL-3B and Ovis2-4B, $1536$ for InternVL3-2B) and partition the 765 images
into five classes by $k$-means on the clean activations, clustering each model separately and merging
the three partitions by co-association consensus. The models agree on the partition (ARI $0.73$),
so it reflects structure shared across encoders rather than an artifact of one; inspection of a few
images per cluster yields the class names \emph{residential street}, \emph{open road}, \emph{night drive},
\emph{handheld photo}, and \emph{urban intersection}.

We split the 765 samples 70/30, stratified by class, and fit a linear classifier on the
clean activations of the training split alone, selecting $C \in \{10^{-2},\dots,10^{2}\}$ by 5-fold
cross-validation within it; all numbers are scored on the held-out split. Corruption does not change
which scene an image shows, so a corrupted activation keeps its clean label. Gaussian-AcT matches the
per-dimension mean and standard deviation of the corrupted activations to those of the clean ones; it
is label-free and fitted on all samples of a condition. We report means over five splits $\pm$ SEM
and summarize recovery by the fraction of the clean--corrupted gap closed,
$\rho = (a_{\mathrm{GA}} - a_{\mathrm{none}})/(a_{\mathrm{clean}} - a_{\mathrm{none}})$.

Severity-5 corruption is destructive in pooled projector space: accuracy falls from $94.5\%$ to
$57.6 \pm 0.3\%$, cutting the margin over the $30.4\%$ majority baseline from 64 to 27 points.
Gaussian-AcT lifts it to $84.0 \pm 0.2\%$, recovering $\rho = 0.65$--$0.78$ of the gap. The effect
is uniform rather than an average: Gaussian-AcT improves all 45 model--corruption pairs
(\Cref{tab:per_corruption}), and $\rho$ exceeds $0.6$ for 13 of the 15 corruptions, peaking
at $0.95$ for InternVL3-2B under defocus blur. Recovery is largest where the degradation is largest:
the noise family is both the most destructive without adaptation ($49.6\%$) and the most fully
repaired ($\rho = 0.82$), so the transport does not merely improve conditions that were already
easy. Motion blur ($\rho = 0.36$) and pixelate ($0.51$) recover least, as expected of a transport
with one scale and one offset per channel. Raw, uncentered features change $\rho$ by at most $0.03$.

\begin{table}[t]\centering\scriptsize
\setlength{\tabcolsep}{1.4pt}
\caption{Per-corruption effect of the projector transport, by readout. $\Delta$ is the change in accuracy over no TTA (percentage points) and $\rho=\Delta/(\text{clean}-\text{no TTA})$ the fraction of the corruption gap recovered. \emph{Classification} fits and applies the transport on one pooled vector per image and reads it out with a linear probe on scene pseudo-classes (mean over five stratified splits); \emph{MCQ} fits and applies it on the unpooled token sequence inside the model, shared across positions, and reads out RealWorldQA accuracy. Per-corruption $\Delta$ is a point estimate; the means carry the SEM of the mean.}
\label{tab:per_corruption}
\begin{tabular}{l rr rr @{\hskip 8pt} rr rr @{\hskip 8pt} rr rr}
\toprule
& \multicolumn{4}{c}{InternVL3-2B} & \multicolumn{4}{c}{Qwen2.5-VL-3B} & \multicolumn{4}{c}{Ovis2-4B} \\
\cmidrule(lr){2-5}\cmidrule(lr){6-9}\cmidrule(l){10-13}
& \multicolumn{2}{c}{Class.} & \multicolumn{2}{c}{MCQ} & \multicolumn{2}{c}{Class.} & \multicolumn{2}{c}{MCQ} & \multicolumn{2}{c}{Class.} & \multicolumn{2}{c}{MCQ} \\
\cmidrule(lr){2-3}\cmidrule(lr){4-5}\cmidrule(lr){6-7}\cmidrule(lr){8-9}\cmidrule(lr){10-11}\cmidrule(lr){12-13}
Corruption & \makecell{$\Delta$\\(pp)} & \makecell{$\rho$\\(of gap)} & \makecell{$\Delta$\\(pp)} & \makecell{$\rho$\\(of gap)} & \makecell{$\Delta$\\(pp)} & \makecell{$\rho$\\(of gap)} & \makecell{$\Delta$\\(pp)} & \makecell{$\rho$\\(of gap)} & \makecell{$\Delta$\\(pp)} & \makecell{$\rho$\\(of gap)} & \makecell{$\Delta$\\(pp)} & \makecell{$\rho$\\(of gap)} \\
\midrule
\multicolumn{13}{l}{\emph{Noise}}\\
\quad Gaussian   & $+41.3$ & $+0.89$ & $+0.4$ & $+0.05$ & $+35.9$ & $+0.75$ & $-0.7$ & $-0.04$ & $+33.0$ & $+0.81$ & $-0.3$ & $-0.02$ \\
\quad Shot       & $+39.7$ & $+0.84$ & $-0.5$ & $-0.06$ & $+36.5$ & $+0.75$ & $-0.7$ & $-0.05$ & $+35.8$ & $+0.84$ & $-0.1$ & $-0.01$ \\
\quad Impulse    & $+40.3$ & $+0.89$ & $-2.0$ & $-0.29$ & $+36.3$ & $+0.79$ & $+1.4$ & $+0.10$ & $+31.7$ & $+0.81$ & $-0.8$ & $-0.07$ \\
\multicolumn{13}{l}{\emph{Blur}}\\
\quad Defocus    & $+35.5$ & $+0.95$ & $+0.1$ & $+0.02$ & $+31.5$ & $+0.78$ & $+0.8$ & $+0.07$ & $+21.1$ & $+0.84$ & $+0.4$ & $+0.04$ \\
\quad Glass      & $+30.7$ & $+0.92$ & $-0.1$ & $-0.03$ & $+29.3$ & $+0.76$ & $+0.7$ & $+0.08$ & $+23.0$ & $+0.86$ & $+1.0$ & $+0.11$ \\
\quad Motion     & $+15.0$ & $+0.33$ & $-0.7$ & $-0.04$ & $+17.5$ & $+0.36$ & $-1.4$ & $-0.07$ & $+17.3$ & $+0.41$ & $-2.9$ & $-0.15$ \\
\quad Zoom       & $+30.8$ & $+0.74$ & $-0.1$ & $-0.01$ & $+26.8$ & $+0.59$ & $-0.9$ & $-0.05$ & $+25.8$ & $+0.68$ & $-0.3$ & $-0.01$ \\
\multicolumn{13}{l}{\emph{Weather}}\\
\quad Snow & $+22.4$ & $+0.88$ & $+0.3$ & $+0.03$ & $+28.3$ & $+0.67$ & $+1.2$ & $+0.09$ & $+19.2$ & $+0.80$ & $+0.9$ & $+0.07$ \\
\quad Frost & $+23.6$ & $+0.63$ & $-0.5$ & $-0.07$ & $+29.7$ & $+0.65$ & $+1.7$ & $+0.17$ & $+25.7$ & $+0.60$ & $+0.9$ & $+0.09$ \\
\quad Fog & $+12.6$ & $+0.92$ & $-0.9$ & $-0.29$ & $+31.3$ & $+0.79$ & $+1.2$ & $+0.14$ & $+15.0$ & $+0.79$ & $+0.3$ & $+0.04$ \\
\quad Brightness & $+8.2$ & $+0.76$ & $+0.1$ & $+0.05$ & $+9.4$ & $+0.68$ & $+1.3$ & $+0.25$ & $+4.1$ & $+0.57$ & $-0.1$ & $-0.11$ \\
\multicolumn{13}{l}{\emph{Digital}}\\
\quad Contrast & $+40.8$ & $+0.88$ & $-0.8$ & $-0.08$ & $+36.3$ & $+0.70$ & $+2.7$ & $+0.16$ & $+30.8$ & $+0.80$ & $0.0$ & $0.00$ \\
\quad Elastic    & $+33.6$ & $+0.84$ & $+1.3$ & $+0.19$ & $+30.4$ & $+0.62$ & $-1.6$ & $-0.14$ & $+24.7$ & $+0.77$ & $+0.3$ & $+0.03$ \\
\quad Pixelate & $+30.3$ & $+0.61$ & $+0.7$ & $+0.04$ & $+18.6$ & $+0.35$ & $-0.9$ & $-0.05$ & $+29.3$ & $+0.59$ & $+1.2$ & $+0.07$ \\
\quad JPEG       & $+18.0$ & $+0.91$ & $-0.1$ & $-0.04$ & $+20.7$ & $+0.71$ & $-0.4$ & $-0.06$ & $+9.9$ & $+0.81$ & $-1.0$ & $-0.26$ \\
\midrule
\textbf{Mean} & \lnc{+28.2}{0.3} & $+0.80$ & \lnc{-0.2}{0.5} & $-0.04$ & \lnc{+27.9}{0.3} & $+0.66$ & \lnc{+0.3}{0.5} & $+0.04$ & \lnc{+23.1}{0.2} & $+0.73$ & \lnc{0.0}{0.5} & $-0.01$ \\
\bottomrule
\end{tabular}
\end{table}

This recovery does not reach the model's multiple-choice answers.
Averaged over the corruptions, the probe recovers $\rho = 0.78$, $0.65$, and $0.72$ of its clean--corrupted gap for InternVL3-2B, Qwen2.5-VL-3B, and Ovis2-4B.
For those answers, the same map is fitted offline on the same corrupted images and applied at the same projector output, but to every visual token rather than to one pooled vector per image, and the language model rather than a linear probe reads the result.
Position-shared Gaussian-AcT then recovers $\rho = -0.04$, $0.04$, and $-0.01$ of the RealWorldQA accuracy gap (\Cref{fig:teaser}).
The two measurements thus share the site, the form of the map, and the offline access to the corrupted set; they differ only in what the map acts on and who reads it.

The corrupted representation therefore still contains the scene information, and matching its first two moments makes that information linearly decodable again, yet the answer does not follow.
Restoring the first two moments of the projector distribution is thus not sufficient for the downstream task, and probe accuracy alone overstates what feature alignment delivers.

Two caveats bound these conclusions. The labels are not independent of the features: they are
derived by clustering the clean activations of all 765 samples, so the label definition has seen the
test images and absolute accuracies are inflated; the clean--corrupted drop and the recovered
fraction $\rho$, which carry the conclusions, are unaffected in direction. The probe and its
transport also operate on pooled sample vectors, whereas adaptation acts on the per-token
distribution inside the model, so this experiment diagnoses what the representation retains and
does not itself constitute an adaptation method.

\section{Per-Corruption Decomposition of the Squared Shift}
\label{app:per_corruption}
\begin{figure}[t]
    \centering
    \includegraphics[width=\linewidth]{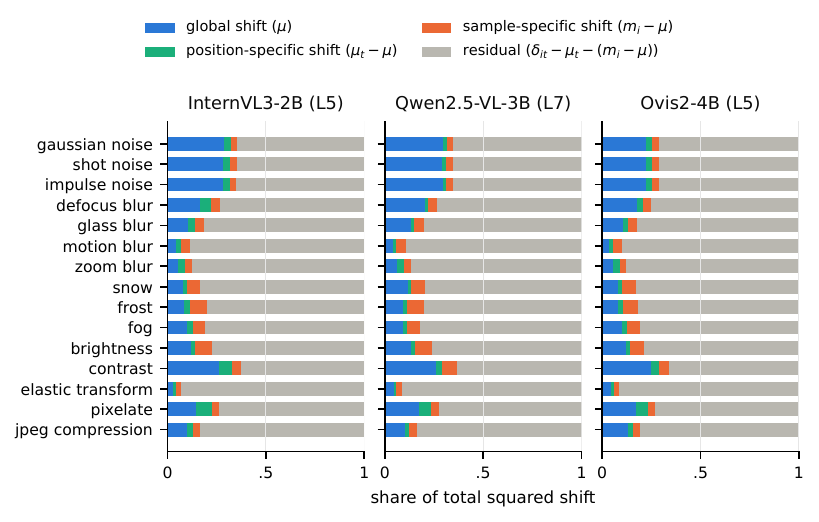}
    \caption{\textbf{Per-corruption decomposition of the total squared shift
    (\Cref{eq:shift_decomposition})} at a shallow block (blocks 5/7/5 for
    InternVL3-2B/Qwen2.5-VL-3B/Ovis2-4B), the most favorable case for sample-blind adaptation: share
    of the total squared shift captured by the global shift $\mu$, the position-specific shift
    $\mu_t-\mu$, the sample-specific shift $m_i-\mu$, and the residual. Only the noise family and
    contrast produce a sizable global component (up to $0.30$); geometric and compression corruptions
    (elastic transform, JPEG, motion blur) are almost entirely residual.}
    \label{fig:decomposition_per_corruption}
\end{figure}

\Cref{fig:decomposition_per_corruption} breaks the decomposition of \Cref{fig:decomposition} down
by corruption at a shallow block, the most favorable case for sample-blind adaptation.

\section{Substituting Clean Activations}
\label{app:substitution}

\subsection{Selecting the substituted subset by importance}
\label{app:selection}
\begin{figure}[t]
\centering
\includegraphics[width=0.85\textwidth]{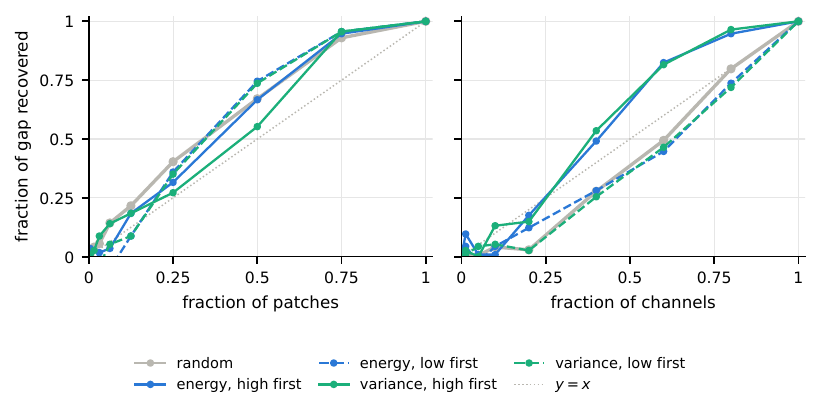}
\caption{\textbf{Importance-based selection does not beat random substitution.} Fraction of the
clean--corrupted gap recovered against the fraction of patches (left) or channels (right) whose
clean activation is substituted, choosing the subset at random or by ranking units on the
magnitude or on the across-dataset variance of their clean activations, highest or lowest first;
Qwen2.5-VL-3B,
Gaussian noise at severity~5, dotted line $y=x$.}
\label{fig:interpolation_frac_patches}
\end{figure}

To test whether the recovery curves of \Cref{sec:substitution} reflect the structure of the
representation or only the statistics of random sampling, we repeated the sweep with the
substituted subset chosen by importance rather than at random. We rank patches and channels
either by the magnitude of their clean activations or by how much these activations vary across the
dataset, two criteria that use no knowledge of where the corruption acts, and substitute either the
highest- or the lowest-ranked units first. No rule changes the shape of the curve
(\Cref{fig:interpolation_frac_patches}): each retains the profile of random selection, including the
flat region at small subset sizes, and differs from it by a roughly constant horizontal factor. Only
the highest-ranked channel rules improve on random selection; the lowest-ranked rules match it, and
on the patch axis informed selection is never better than random and often slightly worse. Choosing
which units to restore therefore rescales the recovery curve at best: the corruption is spread
broadly enough that no cheaply identifiable subset carries the recovery.

\subsection{Additional models}
\label{app:substitution_models}
\begin{figure}[t]
\centering
\begin{minipage}[t]{0.48\textwidth}
  \centering
  \includegraphics[width=0.9\linewidth]{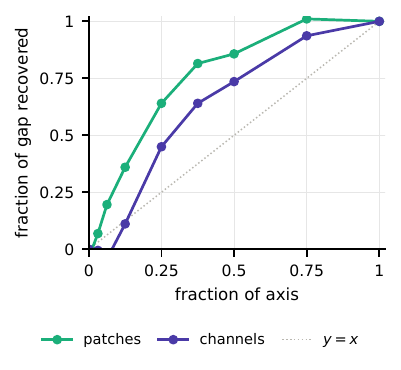}
  \captionsetup{width=\linewidth}
  \captionof{figure}{\textbf{No small subset of the representation carries the degradation.} The fraction of the clean--corrupted gap
    recovered on both axes (dotted: proportional), for InternVL3-2B under Gaussian noise at
    severity~5.}
  \label{fig:substitute_neurons_patches_intern3}
\end{minipage}
\hfill
\begin{minipage}[t]{0.48\textwidth}
  \centering
  \includegraphics[width=0.9\linewidth]{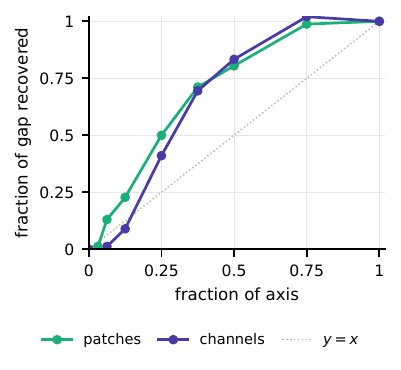}
  \captionsetup{width=\linewidth}
  \captionof{figure}{\textbf{No small subset of the representation carries the degradation.} The fraction of the clean--corrupted gap
    recovered on both axes (dotted: proportional), for Ovis2-4B under Gaussian noise at
    severity~5.}
  \label{fig:substitute_neurons_patches_ovis2}
\end{minipage}
\end{figure}

\Cref{fig:substitute_neurons_patches_intern3,fig:substitute_neurons_patches_ovis2} repeat the random substitution sweep of \Cref{sec:substitution} for InternVL3-2B and Ovis2-4B.

\section{Alternatives to Feature Alignment}
\label{app:alternatives}

\subsection{Prompting does not mitigate image corruptions}
\label{app:prompting}

A natural objection to correcting activations is that the same robustness might be obtained by simply telling the model that its input is corrupted. We test this directly. The evaluation follows \Cref{sec:setup} with a single change, the system prompt: against a baseline that leaves each model's default system message unchanged, we evaluate a variant that appends
\begin{quote}\small
\texttt{The image may be degraded by visual artifacts that carry no information about the scene. Do not describe or reason about the artifacts. Recover the underlying content from large-scale evidence - object shape, size, position, and context - rather than fine local texture, and answer as if you had been given a clean, sharp version of the same image.}
\end{quote}

\begin{figure}[t]
  \centering
  \includegraphics[width=2.7in]{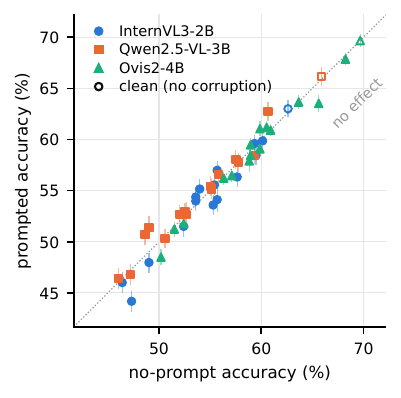}
  \caption{\textbf{Prompting does not recover corrupted accuracy.} One point per model and condition; hollow markers are the clean cells. Error bars are the paired (McNemar) standard error of the difference, so a bar crossing the identity line is consistent with no effect.}
  \label{fig:prompt_null}
\end{figure}

\Cref{fig:prompt_null} reports both variants. The prompt has no measurable effect: averaged over the 15 corruptions, it changes accuracy by $-0.5$, $+0.5$, and $-0.4$~pp for InternVL3-2B, Qwen2.5-VL-3B, and Ovis2-4B, respectively, and no individual corruption differs by more than the noise of the paired comparison: four of the 45 corrupted cells reach nominal $p<0.05$ under a paired sign test, against $2.25$ expected by chance at the $0.05$ level, and three of the four are negative. Clean accuracy is unchanged as well ($+0.4$, $+0.3$, and $+0.0$~pp), so the prompt neither improves accuracy under corruption nor trades clean accuracy for robustness. The signs are inconsistent across models and corruption families, and the prompted and unprompted variants produce nearly identical predictions --- the median cell leaves only 48 of 765 answers discordant; we therefore read the result as a null effect rather than as weak evidence of a benefit.

\subsection{Confidence is a weak predictor of correctness}
\label{app:entropy}
\begin{figure}[t]
  \centering
  \includegraphics[width=0.42\textwidth]{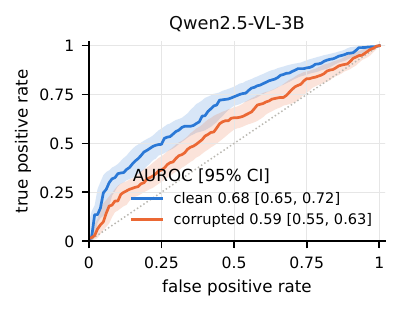}
  \caption{\textbf{Confidence is a weak predictor of correctness.} ROC curves of the model's
  confidence in its answer as a predictor of correctness, for Qwen2.5-VL-3B on RealWorldQA, clean
  and under Gaussian noise at severity~5; shaded bands are 95\% confidence intervals.}
  \label{fig:auroc}
\end{figure}

Entropy- and confidence-based TTA rests on the assumption that a model is confident when it is right and uncertain when it is wrong; if so, making predictions more confident also makes them more correct. We test the assumption rather than the method: instead of running an entropy-minimization procedure, we measure, for Qwen2.5-VL-3B on RealWorldQA, how well the model's confidence in its answer predicts whether the answer is correct, as the area under the ROC curve (AUROC), where $0.5$ means that confidence carries no information about correctness. On clean images the signal is modest (AUROC $0.684$, $95\%$ CI $[0.645, 0.720]$); under Gaussian noise it falls to $0.588$ $[0.549, 0.628]$, barely above the chance band of $[0.460, 0.541]$ expected at this sample size. The assumption fails where it matters most: on all $168$ questions that corruption turns from correct to incorrect, the model's top-ranked answer is a distractor, held with an average margin of $0.258$ over the correct option, so a confidence-increasing objective can only entrench the predictions that would need to be overturned.

This mismatch is structural rather than incidental. Entropy-based TTA was developed for closed-set classification, where each logit corresponds to a fixed semantic class: the ``dog'' direction means the same thing for every input, so a stream of test samples provides repeated evidence about the same decision boundaries, and methods can accumulate class prototypes, adapt normalization statistics, or tune prompts that encode class semantics. In multiple-choice VQA no such correspondence exists. The output symbols are option letters whose meaning is defined by the question and by the ordering of the choices, so the token \texttt{A} denotes an unrelated proposition on each item, and the option set itself varies in size across the benchmark. Nothing is shared across samples for adaptation to accumulate: there are no persistent class directions to sharpen toward, and confidence in \texttt{A} on one question is not evidence about \texttt{A} on the next. Prompt-tuning methods are the partial exception, since the tuned prompt is a genuinely shared parameter and is not tied to any particular option letter; what it cannot do is carry class semantics, as it does when a prompt is optimized alongside a fixed label set. Any gain must therefore come from generic effects such as an improved instruction or output format, and it is still driven by a confidence signal that our measurements show is only weakly related to correctness. What remains is per-sample sharpening of a distribution over symbols, which cannot change the answer, and parameter updates guided by that same weak signal.

\end{document}